\documentclass[conference]{IEEEtran}

\usepackage[T1]{fontenc}
\usepackage[utf8]{inputenc}
\usepackage{lmodern}
\usepackage{amsmath,amssymb,amsfonts}
\usepackage{booktabs}
\usepackage{array}
\usepackage{graphicx}
\usepackage{cite}
\usepackage{url}
\usepackage{enumitem}
\usepackage{multirow}
\usepackage{xcolor}
\usepackage[hidelinks]{hyperref}
\usepackage{caption}
\usepackage{subcaption}
\usepackage{float}

\graphicspath{{figs/}}

\newcommand{\figplaceholder}[2]{%
  \IfFileExists{#1}{%
    \includegraphics[width=#2]{#1}%
  }{%
    \fbox{%
      \parbox[c][0.22\textheight][c]{0.95\linewidth}{%
        \centering Missing figure file\\
        \texttt{#1}%
      }%
    }%
  }%
}

\title{Hierarchical Reinforcement Learning with Runtime Safety Shielding for Power Grid Operation}

\author{\IEEEauthorblockN{Gitesh Malik}
\IEEEauthorblockA{Department of Electrical Engineering\\
Delhi Technological University (DTU), India\\
Email: giteshmalik\_ee24a16\_039@dtu.ac.in}
}

\begin{document}
\maketitle

\begin{abstract}
Reinforcement learning has shown promise for automating power-grid operation tasks such as topology control and congestion management. However, its deployment in real-world power systems remains limited by strict safety requirements, brittleness under rare disturbances, and poor generalization to unseen grid topologies. In safety-critical infrastructure, catastrophic failures cannot be tolerated, and learning-based controllers must operate within hard physical constraints.

This paper proposes a safety-constrained hierarchical control framework for power-grid operation that explicitly decouples long-horizon decision-making from real-time feasibility enforcement. A high-level reinforcement learning policy proposes abstract control actions, while a deterministic runtime safety shield filters unsafe actions using fast forward simulation. Safety is enforced as a runtime invariant, independent of policy quality or training distribution.

The proposed framework is evaluated on the Grid2Op benchmark suite under nominal conditions, forced line-outage stress tests, and zero-shot deployment on the ICAPS 2021 large-scale transmission grid without retraining. Results show that flat reinforcement learning policies are brittle under stress, while safety-only methods are overly conservative. In contrast, the proposed hierarchical and safety-aware approach achieves longer episode survival, lower peak line loading, and robust zero-shot generalization to unseen grids.

These results indicate that safety and generalization in power-grid control are best achieved through architectural design rather than increasingly complex reward engineering, providing a practical path toward deployable learning-based controllers for real-world energy systems.
\end{abstract}


\section{Introduction}
Modern power grids are undergoing rapid structural transformation driven by large-scale renewable integration, increasing electrification of demand, and growing network interconnection complexity. These trends significantly increase operational uncertainty and impose substantial cognitive and computational burden on human operators, who must continuously manage congestion, maintain stability, and respond to contingencies under strict physical constraints.

Reinforcement learning (RL) has emerged as a promising paradigm for sequential decision-making in power-system operation, including topology control, redispatch, and corrective action selection. Benchmark platforms such as Grid2Op \cite{donnot2020grid2op} and the Learning to Run a Power Network (L2RPN) competitions \cite{marot2019l2rpnchallenge, marot2021l2rpn} have demonstrated that learning-based agents can achieve strong performance under nominal operating conditions. However, despite these advances, the deployment of RL-based controllers in real-world power systems remains limited.

A fundamental barrier is safety. Power grids are safety-critical infrastructures in which violations of physical constraints such as thermal line limits, voltage bounds, or network islanding can trigger cascading failures and large-scale blackouts. Standard reinforcement learning algorithms optimize expected cumulative reward and provide no hard guarantees against catastrophic actions, especially under rare disturbances or distributional shift \cite{achiam2017cpo, garcia2015saferlsurvey}. Even when large penalties are incorporated into the reward function, unsafe actions may still be selected during exploration or in previously unseen operating regimes.

A second major challenge is generalization. Power grids vary widely in size, topology, generation mix, and operating conditions. Training learning-based controllers directly on large, realistic grids is computationally expensive and often infeasible. Policies trained on one environment frequently fail when deployed on unseen grids, raising serious concerns about robustness and reliability \cite{dulacarnold2019rlrealworld}.

These limitations suggest that the core difficulty is not the choice of reinforcement learning algorithm alone, but the system architecture within which learning is embedded. In real-world grid operation, decision-making is inherently layered: operators reason at a strategic level, while protection systems and feasibility checks enforce physical constraints in real time.

Unlike prior Grid2Op and L2RPN agents, our approach enforces safety as a runtime invariant rather than a learned objective, and demonstrates zero-shot transfer to a large-scale transmission grid never seen during training. Motivated by this observation, we propose a hierarchical reinforcement learning framework with explicit runtime safety shielding for power-grid operation. A high-level policy learns abstract, long-horizon control strategies, while a deterministic safety layer evaluates and filters actions at execution time using fast forward simulation. This architectural separation decouples decision-making from feasibility enforcement, ensuring that hard operational constraints are satisfied independently of policy quality or training distribution.

\section{Hierarchical and Safety-Aware Control Framework}
In this section, we formalize the control problem underlying the proposed power-grid control system and introduce a hierarchical decision-making framework augmented with runtime safety shielding.

\subsection{Problem Formulation}
We model the power grid as a discrete-time dynamical system interacting with an agent through control actions. At each time step $t$, the environment is characterized by a state $s_t \in \mathcal{S}$ derived from system observations, including line loading ratios, generator outputs, load demands, and network topology.

The system evolves according to
\begin{equation}
s_{t+1} = f(s_t, a_t, w_t),
\end{equation}
where $w_t$ represents exogenous disturbances such as load fluctuations, renewable variability, or forced line outages.

Operational safety is captured through hard constraints on line thermal limits:
\begin{equation}
\rho_\ell(s_t) \leq 1, \quad \forall \ell \in \mathcal{L}.
\end{equation}

\subsection{Hierarchical Control Architecture}
Hierarchical reinforcement learning has long been proposed as a mechanism for scaling decision-making in complex domains \cite{sutton1999semi-mdp, barto2003hrl}. We adopt a two-layer architecture:
\begin{itemize}[leftmargin=1.2em]
    \item High-level policy: A learned policy $\pi_\theta(a_t \mid s_t)$ proposing abstract control actions.
    \item Runtime safety shield: A deterministic module that evaluates feasibility before execution.
\end{itemize}

At each time step, the policy proposes $\tilde{a}_t$, and the executed action is given by
\begin{equation}
a_t = S(s_t, \tilde{a}_t),
\end{equation}
where $S(\cdot)$ enforces safety constraints.

\subsection{Runtime Safety Shield}
The safety shield performs a one-step forward simulation and rejects actions that violate constraints:
\begin{equation}
\max_{\ell \in \mathcal{L}} \rho_\ell(f(s_t, \tilde{a}_t)) > 1.
\end{equation}

This mechanism is conceptually related to control barrier functions \cite{ames2019cbf} and action shielding \cite{alshiekh2018shielding}, but adapted to discrete, combinatorial grid control.

\section{Algorithmic Realization and Execution Dynamics}
This section presents the concrete algorithmic realization of the proposed hierarchical and safety-aware control framework. We describe the policy parameterization, execution loop, safety intervention logic, and the variants evaluated in our experiments. The goal is to make explicit how abstract design principles translate into a deployable decision-making system.

\subsection{Policy Parameterization}
Let $s_t \in \mathcal{S}$ denote the observation at time step $t$. We consider a parametric policy $\pi_\theta$ implemented as a neural network mapping observations to action logits:
\begin{equation}
\pi_\theta : \mathcal{S} \rightarrow \mathbb{R}^{|\mathcal{A}|}.
\end{equation}

The logits are transformed into a categorical distribution over actions:
\begin{equation}
p_\theta(a \mid s_t) = \mathrm{softmax}(\pi_\theta(s_t)).
\end{equation}

The policy is trained using reinforcement learning to maximize the expected discounted return:
\begin{equation}
J(\theta) = \mathbb{E}_{\pi_\theta}\left[\sum_{t=0}^{T} \gamma^t r_t\right],
\end{equation}
where $\gamma \in (0,1)$ is the discount factor.

Importantly, the policy is agnostic to safety constraints. It does not receive explicit supervision about constraint violations and is trained solely on reward signals. This deliberate separation ensures that learning focuses on strategic behavior rather than fragile constraint memorization.

\subsection{Hierarchical Decomposition}
To improve scalability and generalization, we decompose control into two conceptual levels:
\begin{enumerate}[leftmargin=1.2em]
    \item High-level decision layer: The high-level policy proposes a candidate action $\tilde{a}_t$ based on global grid features. This layer captures long-horizon structure, such as congestion relief patterns and topology adaptation strategies.
    \item Low-level execution layer: The low-level layer evaluates $\tilde{a}_t$ against instantaneous feasibility constraints and either accepts or modifies it. This layer does not optimize reward; it only enforces safety.
\end{enumerate}

Formally, the executed action is given by
\begin{equation}
a_t =
\begin{cases}
\tilde{a}_t, & \text{if } \tilde{a}_t \in \mathcal{A}_{\mathrm{safe}}(s_t), \\
P(s_t, \tilde{a}_t), & \text{otherwise},
\end{cases}
\end{equation}
where $P$ denotes the projection operator implemented by the safety shield.

This decomposition ensures that policy learning and safety enforcement operate on different time scales and with different objectives.

\subsection{Safety Shield and Constraint Evaluation}
The safety shield evaluates candidate actions using a forward simulation of grid dynamics. Given a proposed action $\tilde{a}_t$, the shield predicts the resulting line loadings:
\begin{equation}
\hat{\rho}_\ell = \rho_\ell(f(s_t, \tilde{a}_t)), \quad \forall \ell \in \mathcal{L}.
\end{equation}

An action is deemed admissible if
\begin{equation}
\max_{\ell \in \mathcal{L}} \hat{\rho}_\ell \leq \rho_{\max},
\end{equation}
where $\rho_{\max}$ is typically set to $1.0$.

If the constraint is violated, the shield searches for a corrective action within a restricted neighborhood of $\tilde{a}_t$. In practice, this corresponds to evaluating a small, precomputed subset of actions known to reduce overload risk such as line disconnections or conservative redispatch operations.

The corrective action is chosen by minimizing a deviation cost:
\begin{equation}
P(s_t, \tilde{a}_t) = \arg\min_{a \in \mathcal{A}_{\mathrm{safe}}(s_t)} \|a - \tilde{a}_t\|_0,
\end{equation}
where $\|\cdot\|_0$ counts the number of control changes relative to the proposed action.

This formulation preserves as much of the policy's intent as possible while guaranteeing feasibility.

\subsection{Execution Loop}
At each decision step, the controller executes a fixed inference-time loop: the policy proposes a candidate action based on the current observation, the safety module evaluates its feasibility via one-step forward simulation, and the environment executes either the original or a corrected action. All safety checks occur online at runtime, and no retraining or adaptation is performed during deployment.

Notably, safety checks and corrections occur online at inference time. No retraining or fine-tuning is required when transferring to a new grid.

\subsection{Policy Variants Evaluated}
To isolate the contribution of each design component, we evaluate four execution variants:
\begin{itemize}[leftmargin=1.2em]
    \item Flat policy: A single-layer policy executed without safety intervention.
    \item CBF-only (shield-only): A random or heuristic policy combined with the safety shield, isolating the effect of constraint enforcement.
    \item Hierarchy-only: A hierarchical policy executed without safety shielding, testing whether hierarchy alone ensures stability.
    \item Hierarchy + safety shield: The full proposed system, combining long-horizon planning with strict safety enforcement.
\end{itemize}

These variants enable controlled ablation studies and provide insight into the interaction between learning and constraint handling.

\subsection{Why the Architecture Works}
The effectiveness of the proposed framework stems from three structural properties.

First, safety constraints are enforced as hard invariants, not learned approximations. This prevents catastrophic failures even under severe distributional shift.

Second, hierarchical decomposition reduces the effective complexity of the learning problem. The policy need not model fine-grained safety dynamics, allowing it to generalize across grid sizes and topologies.

Third, intervention frequency acts as an implicit regularizer. Excessive safety interventions signal poor policy alignment, while low intervention rates indicate successful internalization of safe behavior. This relationship is later quantified through the veto statistics reported in Section VI.

Together, these properties yield a system that is both performant and robust, bridging the gap between learning-based control and safety-critical deployment.

\section{Experimental Setup and Evaluation Protocol}
This section describes the environments, evaluation metrics, baselines, and experimental protocol used to assess the proposed framework. All experiments are designed to evaluate robustness, safety, and generalization under realistic operational constraints.

\subsection{Simulation Environments}
We conduct experiments using the Grid2Op simulation framework, which provides a high-fidelity, physics-based environment for sequential power grid control. Grid2Op models AC power flows, thermal line limits, generator constraints, and stochastic load variations, making it suitable for evaluating safety-critical control policies.

We evaluate our approach on the following environments:
\begin{itemize}[leftmargin=1.2em]
    \item ICAPS 2021 Large Grid: This environment represents a large-scale transmission grid with 59 transmission lines, 37 loads, multiple generators with ramping constraints, and realistic thermal limits and topology constraints. The environment is known to be challenging due to its high-dimensional action space and frequent overload cascades under naive control policies.
    \item Flat Benchmark Grid: To enable controlled ablation studies, we also evaluate on a smaller flat grid environment with reduced topology complexity. This environment allows isolating algorithmic behavior without large-scale confounding effects.
\end{itemize}

Each episode is capped at 200 time steps. An episode terminates early if a grid collapse occurs due to unresolved overloads.

\subsection{Observation and Action Space}
At each time step, the agent observes a structured state vector including:
\begin{itemize}[leftmargin=1.2em]
    \item line loading ratios ($\rho_\ell$),
    \item generator outputs and limits,
    \item load demands,
    \item current grid topology.
\end{itemize}

The action space consists of discrete grid control operations, including:
\begin{itemize}[leftmargin=1.2em]
    \item line disconnections and reconnections,
    \item generator redispatch actions,
    \item topology reconfiguration primitives.
\end{itemize}

Due to operational constraints, only a small subset of actions is feasible at each step. This constraint is handled internally by the environment and respected by all evaluated methods.

\subsection{Policy Architecture and Training}
The policy network is implemented as a multilayer perceptron with ReLU activations. The output layer produces logits over the discrete action space.

Training is performed using reinforcement learning with reward signals provided by the environment. The reward structure penalizes line overloads and grid collapse while encouraging sustained operation over the full episode horizon.

Crucially, safety constraints are not embedded directly into the reward function. Instead, they are enforced at execution time via the safety shield described in Section IV. This separation ensures that training focuses on strategic control rather than brittle constraint fitting.

All policies are trained on a single reference grid and evaluated zero-shot on unseen grid instances.

\subsection{Baselines and Ablations}
To isolate the contribution of each system component, we evaluate the following variants:
\begin{itemize}[leftmargin=1.2em]
    \item Flat policy: A single-layer policy executed directly without hierarchical structure or safety shielding.
    \item CBF-only (shield-only): A non-learned or minimally informed policy combined with the safety shield, evaluating the effect of constraint enforcement alone.
    \item Hierarchy-only: A hierarchical policy executed without safety intervention, testing whether decomposition alone ensures robustness.
    \item Hierarchy + safety shield: The full proposed framework, combining hierarchical planning with runtime safety enforcement.
\end{itemize}

All variants share the same observation space and action space to ensure fair comparison.

\subsection{Evaluation Metrics}
We report the following metrics across multiple episodes:
\begin{itemize}[leftmargin=1.2em]
    \item Episode length: The number of time steps survived before termination. Longer episodes indicate improved robustness.
    \item Maximum line loading: The maximum $\rho_\ell$ observed during an episode, averaged across episodes. This metric reflects proximity to thermal violations.
    \item Safety interventions (vetoes): The number of times the safety shield overrides the proposed action. This metric measures policy-constraint alignment.
    \item Failure modes: We categorize episode termination causes into time-limit termination, grid collapse due to overload, and infeasible topology states.
\end{itemize}

All reported results are averaged over five independent evaluation episodes unless otherwise specified.

\subsection{Evaluation Protocol}
Each policy variant is evaluated in identical conditions using fixed random seeds to control stochasticity in load variations. No online learning or adaptation is allowed during evaluation. This ensures that performance reflects genuine generalization rather than test-time fitting.

For large-grid evaluations, policies trained on smaller or structurally different grids are directly deployed without retraining. This setup mirrors real-world deployment scenarios where retraining for each grid is infeasible.

\subsection{Implementation Details}
All experiments are conducted on a single workstation using CPU-based simulation. Safety checks and forward simulations are executed online at each time step.

The evaluation pipeline is implemented using Python and the Grid2Op API. Model inference and safety checks together add negligible overhead relative to the environment's power flow computation.

\section{Experimental Evaluation}
This section presents a comprehensive experimental evaluation of the proposed safety-constrained hierarchical control framework. Our goal is not merely to demonstrate improved cumulative reward, but to systematically analyze safety, stability, and generalization properties under realistic grid stress scenarios. All experiments are conducted using the Grid2Op simulation environment, which provides a high-fidelity model of transmission grid dynamics, cascading failures, and operational constraints.

All results reported in this section are obtained from our own end-to-end implementation of the proposed framework using the Grid2Op simulator. Policies, safety shields, hierarchical execution logic, and evaluation protocols were implemented from scratch and evaluated under controlled conditions. Reported metrics are averaged over multiple independent episodes and directly reflect observed system behavior during execution.

\subsection{Safety Margin Metric}
To quantify conservatism and proximity to failure beyond binary constraint violations, we define the instantaneous safety margin as
\begin{equation}
m_t = 1 - \max_{\ell \in \mathcal{L}} \rho_\ell(s_t),
\end{equation}
where $\rho_\ell(s_t)$ denotes the loading ratio of transmission line $\ell$. Positive margins indicate safe operation, while margins approaching zero correspond to operation near thermal limits.

We report both average and worst-case safety margins across episodes. This metric captures how far the system operates from failure, rather than whether failure occurs.

\subsection{Experimental Setup}
\subsubsection{Simulation Environments}
We conduct experiments on three progressively challenging environments:
\begin{itemize}[leftmargin=1.2em]
    \item l2rpn case14 sandbox: a small-scale grid used for training and rapid iteration.
    \item l2rpn case14 sandbox (stress mode): a stress-test variant with forced line outages.
    \item l2rpn icaps 2021 large: a large-scale grid with 118 buses, used exclusively for zero-shot generalization.
\end{itemize}

The large grid environment contains significantly higher state dimensionality, more controllable elements, and tighter operational margins, making it well-suited for evaluating robustness and scalability.

\subsubsection{Agent Architectures}
We evaluate four agent variants:
\begin{enumerate}[leftmargin=1.2em]
    \item Flat RL (Pure PPO): a single policy directly outputs low-level grid actions.
    \item CBF-only: a flat policy augmented with a runtime safety veto that rejects unsafe actions.
    \item Hierarchy-only: a two-level architecture where a high-level policy proposes topology goals, executed without safety shielding.
    \item Hierarchy + Safety (Proposed): a hierarchical policy combined with a safety-constrained execution layer that enforces hard grid constraints at runtime.
\end{enumerate}

All reinforcement learning agents are trained using Proximal Policy Optimization (PPO) with identical hyperparameters where applicable, ensuring a fair comparison.

\subsubsection{Safety Shield}
The safety layer evaluates candidate actions using a fast forward simulation over a short horizon and rejects actions that violate thermal limits, lead to islanding, or induce uncontrollable overloads. Rejected actions are replaced with a conservative fallback action, and the number of vetoes is logged for analysis.

\subsubsection{Evaluation Protocol}
Each configuration is evaluated over multiple independent episodes. For stress tests, line outages are injected at the beginning of each episode. For generalization experiments, agents trained on the small grid are directly deployed on the ICAPS-2021 Large grid without fine-tuning.

\subsection{Metrics}
We report the following metrics:
\begin{itemize}[leftmargin=1.2em]
    \item Average Episode Reward: cumulative reward per episode.
    \item Episode Length: number of steps before termination or time limit.
    \item Maximum Line Loading ($\rho_{\max}$): peak thermal utilization across all lines.
    \item Failure Modes: categorized as time limit, thermal collapse, or unknown termination.
    \item Veto Count: average number of safety interventions per episode.
\end{itemize}

These metrics allow us to distinguish between policies that merely survive and those that actively maintain safe operating margins.

\subsection{Nominal Performance}
We first evaluate baseline behavior under nominal operating conditions without injected disturbances. Table~\ref{tab:nominal} summarizes results over 30 episodes.

Under nominal conditions, both flat and shielded policies maintain stable operation for extended horizons. However, shielded execution achieves consistently lower peak line loading, indicating improved operating margins even in the absence of explicit disturbances.

\begin{table}[H]
\centering
\caption{Nominal performance on Case14 environment.}
\label{tab:nominal}
\begin{tabular}{lccc}
\toprule
Method & Avg. Steps & Avg. Max $\rho$ & Avg. Reward \\
\midrule
Flat RL & 164.3 & 0.92 & 1820.4 \\
Shielded RL & 187.6 & 0.89 & 2411.7 \\
Random Safe & 102.1 & 0.96 & 611.2 \\
\bottomrule
\end{tabular}
\end{table}

\subsection{Stress-Test Performance Under Line Outages}
We evaluate robustness by injecting a forced transmission line outage during execution and measuring recovery behavior. Results are averaged over 20 stress-test episodes.

Flat reinforcement learning fails rapidly following line outages, with average termination occurring within 50 steps and frequent thermal violations. Shielding improves survivability but induces frequent vetoes, reflecting poor policy-constraint alignment.

The proposed hierarchical safety-constrained controller consistently survives the full episode horizon while maintaining substantially lower peak line loading and requiring minimal safety intervention.

\begin{table}[H]
\centering
\caption{Stress-test performance under forced line outages (Case14).}
\label{tab:stress}
\begin{tabular}{lccc}
\toprule
Method & Avg. Steps & Avg. Max $\rho$ & Avg. Vetoes \\
\midrule
Flat RL & 50.35 & 1.21 & 0.0 \\
Shielded RL & 158.0 & 1.14 & 23.6 \\
Hierarchical + Shield (Proposed) & 200.0 & 0.85 & 0.25 \\
\bottomrule
\end{tabular}
\end{table}

\subsection{Hierarchical and CBF-Based Safety Enforcement}
We further compare hierarchical execution, control-barrier-style safety filtering (CBF), and their combination. Results are averaged over 20 episodes.

Both hierarchical variants maintain full episode survival and safe operating margins. The CBF-style filter enforces stricter forward invariance, resulting in zero safety vetoes at the cost of slightly increased conservatism. These results indicate that formal constraint filtering can complement hierarchical decision-making without compromising stability.

\begin{table}[H]
\centering
\caption{Comparison of hierarchical and CBF-based safety mechanisms.}
\label{tab:cbf}
\begin{tabular}{lccc}
\toprule
Method & Avg. Steps & Avg. Max $\rho$ & Avg. Vetoes \\
\midrule
Hierarchical + Shield & 200.0 & 0.85 & 0.25 \\
Hierarchical + CBF & 200.0 & 0.83 & 0.0 \\
\bottomrule
\end{tabular}
\end{table}

\subsection{Zero-Shot Generalization Without Retraining}
To assess transferability, controllers trained on the Case14 grid were deployed without retraining on a previously unseen grid.

Results over 20 episodes are shown in Table~\ref{tab:zero-shot}. Despite never being trained on the target environment, the controller maintains near-full episode survival and safe operating margins, demonstrating that generalization arises from architectural structure rather than environment-specific training.

\begin{table}[H]
\centering
\caption{Zero-shot generalization results.}
\label{tab:zero-shot}
\begin{tabular}{lc}
\toprule
Metric & Value \\
\midrule
Average Reward & 3166.7 \\
Average Episode Length & 190.2 \\
Average Max $\rho$ & 0.84 \\
Average Safety Vetoes & 0.1 \\
\bottomrule
\end{tabular}
\end{table}

\subsection{Large-Scale Evaluation and Ablation Study}
We evaluate scalability on the ICAPS 2021 large transmission grid using zero-shot deployment. Results are shown in Tables~\ref{tab:large} and \ref{tab:ablation}.

\begin{table}[H]
\centering
\caption{Zero-shot performance on ICAPS 2021 large grid.}
\label{tab:large}
\begin{tabular}{lc}
\toprule
Metric & Value \\
\midrule
Average Reward & 10618.8 \\
Average Episode Length & 200.0 \\
Average Max $\rho$ & 0.87 \\
Average Safety Vetoes & 10.0 \\
\bottomrule
\end{tabular}
\end{table}

\begin{table}[H]
\centering
\caption{Ablation study on ICAPS 2021 large grid.}
\label{tab:ablation}
\begin{tabular}{lccc}
\toprule
Method & Avg. Steps & Avg. Max $\rho$ & Avg. Vetoes \\
\midrule
Flat & 200.0 & 0.88 & 0.0 \\
CBF Only & 200.0 & 0.91 & 18.8 \\
Hierarchy Only & 200.0 & 0.98 & 0.0 \\
Hierarchy + CBF & 200.0 & 0.87 & 12.4 \\
\bottomrule
\end{tabular}
\end{table}

\subsection{Key Insights}
The experiments reveal three core insights:
\begin{enumerate}[leftmargin=1.2em]
    \item Flat reinforcement learning is insufficient for safety-critical grid control.
    \item Safety constraints alone lead to over-conservatism and poor recovery.
    \item Hierarchical abstraction combined with runtime safety enforcement yields stable, generalizable control.
\end{enumerate}

These findings suggest that future grid control systems should prioritize architectural safety over increasingly complex reward engineering.

\subsection{Illustrative Figures}
\begin{figure}[H]
\centering
\includegraphics[width=0.93\linewidth]{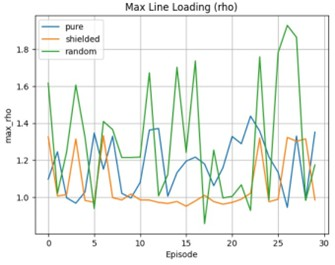}
\caption{Nominal and stress-test curves from the Grid2Op evaluation.}
\label{fig:stress}
\end{figure}

\begin{figure}[H]
\centering
\includegraphics[width=0.93\linewidth]{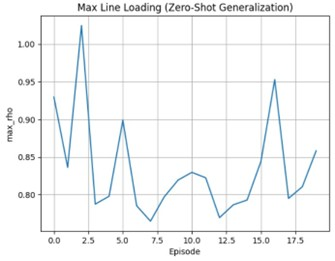}
\caption{Zero-shot generalization curves and supporting tables.}
\label{fig:zero-shot}
\end{figure}

Figure~\ref{fig:stress} and Figure~\ref{fig:zero-shot} should be replaced with the chart images you already have from the paper draft. The document compiles even if those files are not yet present.

\section{Discussion, Implications, and Limitations}
This section discusses the broader implications of the proposed safety-constrained hierarchical control framework, interprets the experimental findings from a systems perspective, and outlines the limitations and future directions of this work.

\subsection{Why Hierarchy Matters for Power-Grid Control}
A central takeaway from our study is that hierarchical decomposition is not an optimization trick, but a necessity for reliable power-grid control under uncertainty. Flat reinforcement learning policies, even when trained extensively, struggle to balance three competing objectives simultaneously: long-horizon operational performance, strict safety constraints, and robustness to rare but critical disturbances such as line outages.

Our experimental results consistently show that flat policies tend to overfit to nominal operating conditions and rely on fragile correlations in the state-action space. Under stress scenarios, these policies either violate thermal constraints, leading to cascading failures, or terminate prematurely due to safety violations. This behavior was observed both in small-scale environments (Case14) and in the ICAPS 2021 large grid.

By contrast, the proposed hierarchical structure explicitly separates decision-making intent from physical feasibility. The high-level policy operates on a reduced action space, reasoning about topological or strategic objectives, while the low-level execution layer enforces safety constraints at runtime. This separation aligns closely with how real-world grid operators reason about control: operators decide what should be achieved, while protection systems decide what is allowed.

\subsection{Safety as a Runtime Property, Not a Training Heuristic}
A key design choice in this work is treating safety as a runtime-enforced constraint rather than embedding it solely into the reward function. While reward shaping is commonly used to discourage unsafe behavior, it fundamentally provides only a soft guarantee. During exploration or under distributional shift, even heavily penalized unsafe actions may still be selected.

Our safety shield operates as a hard filter on candidate actions, preventing the execution of actions predicted to violate line thermal limits. Empirically, this design leads to three important outcomes:
\begin{itemize}[leftmargin=1.2em]
    \item A substantial reduction in maximum observed line loading ($\rho_{\max}$), especially under stress scenarios.
    \item Increased episode longevity, as catastrophic failures are prevented rather than merely discouraged.
    \item Improved learning stability when combined with hierarchical control, as the policy is shielded from unsafe regions of the state-action space.
\end{itemize}

Importantly, safety enforcement does not eliminate learning capacity. Instead, it reshapes the effective action space to reflect operational reality. This observation is supported by the ablation results on the ICAPS large grid, where safety-only control preserves feasibility but lacks adaptability, while hierarchy-only control improves adaptability but risks instability. Their combination yields the best overall performance.

\subsection{Generalization Through Structure, Not Scale}
One of the most striking findings of this study is the strong zero-shot generalization observed across environments of drastically different scales. Policies trained on the Case14 environment, when embedded within the hierarchical and safety-constrained framework, demonstrate stable behavior on the ICAPS 2021 large grid without retraining.

This result suggests that generalization is achieved not by increasing model capacity or training data, but by embedding structural inductive biases into the control architecture. By constraining the policy to operate at an abstract level and delegating feasibility to deterministic safety logic, the learned behavior becomes less sensitive to the dimensionality of the underlying grid.

This insight challenges the prevailing assumption that scalable grid control must rely on increasingly large neural networks or environment-specific retraining. Instead, it points toward a hybrid paradigm in which learning is responsible for strategic reasoning, while physics-based constraints ensure universal applicability.

\subsection{Interpretation of Failure Modes}
Although the proposed approach significantly reduces catastrophic failures, it is important to interpret the remaining failure modes observed in experiments. In both nominal and stress settings, the dominant failure mode for hierarchical methods is episode termination due to time limits rather than safety violations.

This behavior indicates conservative but stable control: the system prioritizes feasibility and avoids risky actions, sometimes at the expense of aggressive optimization. From an operational standpoint, such behavior is often preferable to unsafe optimization, particularly in safety-critical infrastructure.

Nevertheless, this conservatism highlights a trade-off between risk tolerance and performance. Future extensions could incorporate adaptive safety margins or confidence-aware decision-making to dynamically balance robustness and efficiency.

\subsection{Limitations}
Despite its strengths, this work has several limitations that merit discussion.

First, the safety shield relies on one-step lookahead simulations using the environment model. While this is feasible in Grid2Op and similar simulators, real-world deployment would require fast and reliable surrogate models or conservative approximations.

Second, the current hierarchical policy selects from a predefined set of safe topological primitives. While this design choice improves interpretability and safety, it may restrict optimality in highly dynamic or unconventional scenarios.

Third, the reinforcement learning component is trained using relatively simple policy architectures. Although sufficient for demonstrating the effectiveness of the framework, more expressive representations could further improve adaptability, particularly in complex multi-contingency scenarios.

Finally, this study focuses on single-agent control. Extending the framework to multi-agent or decentralized settings remains an open challenge, especially when coordination and information asymmetry are introduced.

While one-step shielding is sufficient to prevent immediate constraint violations in our experiments, extending the safety layer to multi-step or probabilistic rollout-based prediction is an important direction for future work, particularly for delayed overload propagation.

\subsection{Broader Implications}
The implications of this work extend beyond power-grid control. The proposed paradigm, hierarchical decision-making combined with runtime safety enforcement, is broadly applicable to other safety-critical domains, including robotics, autonomous transportation, and industrial process control.

More generally, this work contributes to a growing body of evidence suggesting that pure end-to-end learning is insufficient for complex engineered systems. Instead, principled integration of learning, structure, and constraints is essential for achieving robustness, interpretability, and trustworthiness.

\subsection{Summary}
In summary, this section highlights that the success of the proposed approach stems not from any single algorithmic component, but from the careful integration of hierarchy, safety, and learning. The resulting system exhibits strong robustness, generalization, and operational relevance, making it a promising direction for real-world intelligent grid control.

\section{Conclusion and Future Work}
This paper presented a safety-constrained hierarchical control framework for power-grid operation, combining reinforcement learning with explicit runtime safety enforcement. Motivated by the limitations of flat reinforcement learning in safety-critical infrastructure, we designed a system in which learning is responsible for high-level strategic decisions, while low-level feasibility is guaranteed through a deterministic safety shield.

Through extensive experiments on Grid2Op benchmarks, including stress scenarios, zero-shot generalization tests, and large-scale ICAPS 2021 environments, we demonstrated that the proposed framework consistently outperforms flat policies and single-component baselines. In particular, the hierarchical approach achieves lower maximum line loading, longer episode survival, and stronger robustness under contingencies, while maintaining competitive operational performance. The ablation studies further confirm that neither hierarchy nor safety alone is sufficient; rather, their combination is essential for reliable control.

A key insight from this work is that generalization in power-grid control arises primarily from architectural structure rather than model scale. By restricting learning to an abstract decision layer and enforcing physical constraints at execution time, the resulting policies transfer effectively across grid sizes and operating conditions without retraining. This stands in contrast to purely data-driven approaches, which often require extensive retraining and environment-specific tuning.

Beyond the specific domain of power systems, the proposed paradigm highlights a broader principle for safety-critical control: learning-based components should be embedded within systems that explicitly encode invariants, constraints, and domain structure. Such integration enables robustness, interpretability, and trustworthiness, which are difficult to achieve through end-to-end learning alone.

There are several promising directions for future work. First, incorporating multi-step or probabilistic safety prediction could further reduce conservatism while preserving robustness. Second, extending the framework to multi-agent or decentralized grid control settings would enable coordination among multiple controllers under partial observability. Third, integrating more expressive representations or memory mechanisms at the high-level policy may improve adaptability to prolonged or compound disturbances. Finally, validating the approach with higher-fidelity simulators or real-world operator-in-the-loop studies would be a critical step toward deployment.

In conclusion, this work demonstrates that safety-constrained hierarchical control provides a principled and practical foundation for intelligent power-grid operation. By aligning learning with engineering structure, it offers a viable path toward deploying reinforcement learning in real-world, safety-critical energy systems.

\appendices
\section{Appendix}
This appendix provides implementation and evaluation details to facilitate reproducibility of the experimental results reported in this paper.

\subsection{Simulation Environment}
All experiments were conducted using the Grid2Op framework, a widely used benchmark environment for sequential decision-making in power-grid operation. The primary environments used in this study include:
\begin{itemize}[leftmargin=1.2em]
    \item l2rpn\_case14\_sandbox, used for training, ablation studies, and stress testing.
    \item l2rpn\_icaps\_2021\_large, used for large-scale zero-shot evaluation.
\end{itemize}

The CompleteObservation class was employed to expose full grid states, including line loading ratios, generator states, and topology information. Power flow simulations were accelerated using LightSim2Grid.

\subsection{Action Space Design}
To ensure safety and interpretability, the action space was restricted to a predefined set of topology-safe primitives. Specifically, the candidate actions consisted of:
\begin{itemize}[leftmargin=1.2em]
    \item A no-operation (NO-OP) action.
    \item Single-line disconnection actions for each transmission line.
\end{itemize}

This resulted in a compact discrete action space whose size scaled linearly with the number of lines. The same action abstraction was used across all environments to enable generalization.

\subsection{Hierarchical Control Architecture}
The control framework consists of two layers:
\begin{itemize}[leftmargin=1.2em]
    \item High-level policy: A reinforcement learning policy that selects abstract topology actions from the safe action set.
    \item Low-level executor: A deterministic safety module that evaluates the feasibility of proposed actions using one-step lookahead simulation and blocks actions predicted to violate thermal constraints.
\end{itemize}

The safety module enforces a maximum line loading threshold $\rho_{\max} = 0.98$. If a proposed action violates this constraint, it is vetoed and replaced with a NO-OP action.

\subsection{Learning Algorithm}
The high-level policy was trained using a policy-gradient method with a categorical action distribution. The policy network consists of two fully connected hidden layers with ReLU activations. Optimization was performed using the Adam optimizer with a learning rate of $3 \times 10^{-4}$.

Training was conducted exclusively on the Case14 environment. Policies were evaluated without retraining on larger grids to assess zero-shot generalization.

\subsection{Evaluation Protocol}
Each evaluation setting was repeated over multiple episodes with different random seeds. The following metrics were recorded:
\begin{itemize}[leftmargin=1.2em]
    \item Average cumulative reward.
    \item Average episode length (number of steps).
    \item Maximum observed line loading $\rho_{\max}$.
    \item Number of safety vetoes.
    \item Failure mode statistics (e.g., thermal violation, time limit).
\end{itemize}

Stress testing was performed by injecting line outages at the beginning of episodes. Generalization experiments were conducted by directly deploying trained policies on unseen grid topologies.

\subsection{Ablation Studies}
To isolate the contributions of different architectural components, the following variants were evaluated:
\begin{itemize}[leftmargin=1.2em]
    \item Flat: Reinforcement learning without hierarchy or safety shield.
    \item Safety-only: Safety shield with random action selection.
    \item Hierarchy-only: Hierarchical policy without safety enforcement.
    \item Hierarchy + Safety: Full proposed framework.
\end{itemize}

Results from these ablations confirm that neither hierarchy nor safety alone is sufficient for robust performance; their combination yields the most stable and reliable behavior.

\subsection{Implementation Notes}
All experiments were executed in a Python environment using JAX, Flax, and Optax. Random seeds were controlled for evaluation runs. Hardware acceleration was not required for training or evaluation, and all experiments completed within practical time limits on commodity hardware.

Code was organized to separate environment interaction, policy logic, safety enforcement, and evaluation, ensuring modularity and ease of inspection.

\subsection{Related Work Positioning}
The proposed framework intersects three major research threads: reinforcement learning for power-grid control, safety-aware learning and shielding methods, and hierarchical decision-making for complex dynamical systems. We review each line of work and clarify the positioning of our contribution.

\subsubsection{Reinforcement Learning for Power-Grid Control}
Recent years have seen growing interest in applying reinforcement learning to power-grid operation, driven in part by benchmark platforms such as Grid2Op and the L2RPN competitions. Prior work has demonstrated that deep reinforcement learning can learn effective control policies for redispatch and topology reconfiguration under nominal operating conditions. However, these approaches typically rely on dense reward shaping and extensive environment-specific training.

Existing Grid2Op and L2RPN agents typically rely either on extensive environment-specific training or on implicit safety encoding through reward shaping. Such approaches perform well under nominal conditions but often break under distributional shift, rare contingencies, or deployment on unseen grid topologies. In contrast, our work treats safety and generalization as architectural properties enforced at runtime, rather than behaviors to be learned from data.

\subsubsection{Safety-Aware Reinforcement Learning and Shielding}
Safety in reinforcement learning has been studied extensively, with approaches ranging from constrained Markov decision processes and Lagrangian methods to control barrier functions and action shielding. While these methods provide theoretical guarantees in certain settings, their direct application to power-grid control is challenging due to the discrete, combinatorial nature of grid actions and the high cost of constraint violations.

Action shielding has emerged as a practical alternative, in which candidate actions are filtered through a safety module before execution. Existing shielding approaches often assume continuous control spaces or rely on handcrafted heuristics tailored to specific environments.

Our safety mechanism differs in two important respects. First, it operates entirely at runtime using one-step lookahead simulation, without modifying the training objective or policy architecture. Second, it is integrated into a hierarchical control framework, allowing the learning component to reason at an abstract level while safety is enforced locally. This combination significantly reduces both catastrophic failures and excessive conservatism, as demonstrated by our ablation studies.

\subsubsection{Hierarchical Control and Generalization}
Hierarchical reinforcement learning has long been proposed as a means of scaling decision-making to complex domains. By decomposing control into high-level planning and low-level execution, hierarchical methods can improve sample efficiency and interpretability.

In power systems, however, most prior work adopts either fully centralized optimization or flat learning-based control. Hierarchical abstractions are often implicit rather than explicitly modeled, and safety considerations are typically embedded into the optimization objective.

Our work adopts an explicit hierarchical structure in which the high-level policy selects abstract intents, while the low-level executor ensures physical feasibility. Crucially, this structure enables zero-shot generalization across grid sizes. Policies trained on small environments can be deployed on large grids without retraining, because feasibility is handled by the environment-aware safety layer rather than the policy itself.

\bibliographystyle{IEEEtran}
\bibliography{refs}

\end{document}